\documentclass[letterpaper]{article} 
\usepackage[draft]{aaai25}  
\usepackage{times}  
\usepackage{helvet}  
\usepackage{courier}  
\usepackage[hyphens]{url}  
\usepackage{graphicx} 
\usepackage{natbib}  
\usepackage{caption} 
\usepackage{algorithm}
\usepackage{algorithmic}

\usepackage{amsmath}
\usepackage{bm}
\usepackage{subcaption}
\usepackage{booktabs}

\usepackage{xcolor} 

\DeclareMathOperator{\RN}{Re}

\title{Transported Memory Networks accelerating Computational Fluid Dynamics}
\author{Matthias Schulz\equalcontrib, Gwendal Jouan\equalcontrib, Daniel Berger, Stefan Gavranovic, Dirk Hartmann\footnote{Corresponding author: hartmann.dirk@siemens.com}}
\affiliations{Siemens Digital Industries Software}

\usepackage{bibentry}

\begin{document}

\maketitle

\frenchspacing

\begin{abstract}
In recent years, augmentation of differentiable PDE solvers with neural networks has shown promising results, particularly in fluid simulations. However, most approaches rely on convolutional neural networks and custom solvers operating on Cartesian grids with efficient access to cell data. This particular choice poses challenges for industrial-grade solvers that operate on unstructured meshes, where access is restricted to neighboring cells only. In this work, we address this limitation using a novel architecture, named \textit{Transported Memory Networks}. The architecture draws inspiration from both traditional turbulence models and recurrent neural networks, and it is fully compatible with generic discretizations. Our results show that it is point-wise and statistically comparable to, or improves upon, previous methods in terms of both accuracy and computational efficiency.
\end{abstract}

\section{Introduction}

Computer Aided Engineering (CAE) has a long history in industrial engineering. Multi-physics simulations, including Computational Fluid Dynamics (CFD), are among the most used simulation technologies in CAE \cite{liu2022eighty,Kelsall2022CFD}. Even though technology has evolved exponentially in terms of speed and accuracy \cite{ruede2018research}, they still take excessive amounts of time. While an enormous speedup has been recently obtained by utilizing GPUs \cite{gavranovic2024fastsolvers}, further acceleration is at risk since CFD problems can be simulated mathematically optimally \cite{becker2001optimal,Hackbusch2013multi}. 

With the rapid evolution of machine learning (ML) technologies, there is a unique potential to address this challenge \cite{karniadakis2021physics, weinan2021dawning}. Corresponding methods promise orders of magnitude of acceleration \cite{hutson2020ai} and applications range from weather predictions \cite{lam2023learning}, medical applications \cite{karniadakis2021physics} to industrial use cases \cite{lavin2021simulation}. Methods comprise pure ML methods, such as Operator Learning \cite{kovachki2021neural} or physics-constrained methods \cite{karniadakis2021physics}, as well as hybrid methods \cite{sanderse2024scientific}.

Within this contribution, we focus on a hybrid method, specifically on the \textit{solver-in-the-loop} approach of \citeauthor{um2020solver} (\citeyear{um2020solver}) and \citeauthor{kochkov2021machine}(\citeyear{kochkov2021machine}). In these two approaches, traditional numerical methods are augmented by ML components to increase their speed and/or accuracy. Both approaches demonstrated remarkable performance (using state-of-the-art industrial CFD codes as the baseline), particularly in their generalization capabilities. However, they employ convolutional neural networks (CNNs) which has significant implications for the underlying solver architecture and thus limits compatibility with large-scale industrial CFD codes.

\begin{figure}[H]
    \centering
    \includegraphics[width=0.98\columnwidth]{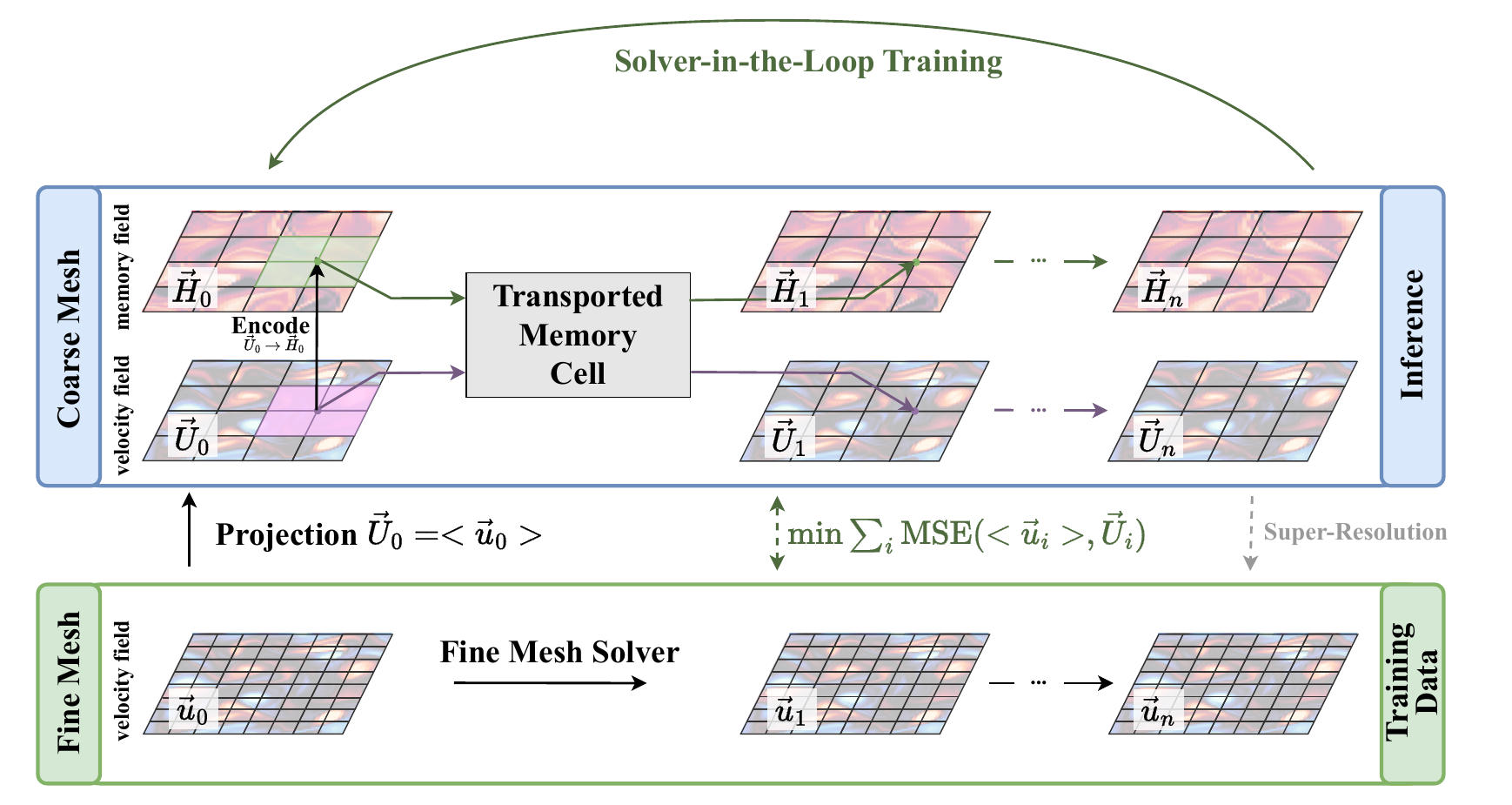}
    \caption{Hybrid \textit{solver-in-the-loop} approach which additionally to physics variables introduces long term memory states. These memory states are transported along the fluid flow as the physics variables are. The corresponding evolution is handled by the Transported Memory Cell (see Figure \ref{fig:cell}), which combines a classical base physics solver with a ML-based augmentation allowing to coarse grain discretiziation without loss of prediction accuracy. }
    \label{fig:overview}
\end{figure}

Our contributions in this work are as follows:
\begin{itemize}
  \item Investigate the impact of the CNN architecture on performance, particularly focusing on the input stencil size.
  \item Propose an alternative approach that utilizes more local information (direct cell neighbors only, making it suitable for state-of-the-art industrial solvers.
  \item Benchmark and evaluate the new approach in comparison to previous approaches.
\end{itemize}

We thereby introduce a new neural network architecture, which we coin \textit{Transported Memory Network}. This architecture leverages the concept of a long term memory (similar to Long Short Term Memory architectures; \citeauthor{hochreiter1997long}, \citeyear{hochreiter1997long}) which is effectively transported with the flow. It thus appropriately reflects the corresponding physics behavior in an Eulerian coordinate systems which is fixed in space rather than advected with the flow. In addition to reflecting physics more correctly, we also expect a better generalization than CNNs when considering more complex geometries. The proposed approach relies only on direct neighbours and is thus relatively geometry agnostic. 


\begin{figure}[H]
    \centering
    \includegraphics[width=0.99\columnwidth]{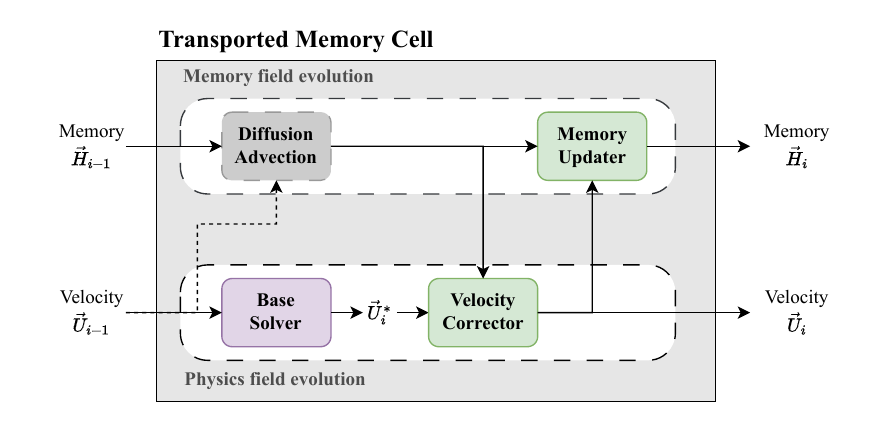}
    \caption{Schematic sketch of the architecture of the transported memory cell, combining a classical base physics solver with a ML-based augmentation. For the specific neural network architectures see Figure \ref{fig:stencil_arch} and \ref{fig:nets_architecture}.} 
    \label{fig:cell}
\end{figure}

\section{Background}

\subsection{Computational Fluid Dynamics}\label{sec:CFD}

\paragraph{Navier-Stokes Equations:} 
Within this contribution we focus on incompressible fluids. These are described by the incompressible Navier-Stokes equations \cite{acheson1990elementary}:
{\footnotesize
\begin{align}
    \rho \left(\partial_t \vec{u} + \vec{u} \cdot \nabla \vec{u} \right) &=
       - \nabla p + \nabla \cdot \bm{\tau} + \rho \vec{f}, \label{eq:NS01}\\
    \nabla \cdot \vec{u} &= 0,   \label{eq:NS02}
\end{align}}

\noindent
where $\vec{u}$ is the flow velocity, $p$ the pressure, $\bm{\tau}$ the deviatoric stress tensor, $\vec{f}$ an external body force, and $\rho$ the constant fluid density. In its native form, $\bm{\tau}$ is linearly dependent on the strain rate tensor, i.e.,  $\nabla \cdot \bm{\tau} = \mu \Delta \vec{u}$ with dynamic viscosity $\mu$. In most industrial applications, solving Equations (\ref{eq:NS01})-(\ref{eq:NS02}) numerically would require very fine spatial and temporal resolutions. Instead, so-called turbulence models extending the Navier-Stokes equations have been proposed to address this challenge \cite{pope2000turbulent}. These turbulence models introduce appropriate corrections to the stress tensor $\bm{\tau}$ and various options have been proposed \cite{pope2000turbulent}.  

\paragraph{Numerical Solvers:}
A broad toolbox of numerical solvers for CFD have been proposed \cite{Kelsall2022CFD} in the past. These include particle based discretizations, e.g., Smooth Particle Hydrodynamics \cite{onate2011particle} or Lattice Boltzman methods \cite{kruger2017lattice}, global discretization methods, e.g., Spectral methods \cite{shen2011spectral}, as well as local grid based discretization methods, e.g., Finite Element \cite{turek1999efficient}, Finite Volume \cite{moukalled2016finite} or Finite Difference schemes \cite{griebel1998numerical}. Within this contribution we aim to improve state-of-the-art industrial methods usually based on Finite Volume (FV) schemes. Thus, we will focus on local discretization methods where we furthermore restrict ourselves to regular structured grids without loss of generality.

\subsection{Pure Neural CFD Solvers}\label{sec:neuralCFD}
As an alternative to numerical solvers a wide range of neural network based pure ML solvers have been proposed recently. They can be classified into two distinct approaches: Operator Learning Methods and Physics-Constrained Methods.

\paragraph{Operator Learning Methods:} These methods directly learn the physics operator, e.g., learn the iteration from one time step to the next (dynamic CFD) or the mapping from a geometry to the corresponding flow field (stationary CFD). That is, they learn the solution operator (numerical solver). Examples include convolutional methods \cite{guo2016convolutional}, DeepONets \cite{lu2021learning}, Fourier Neural Operators \cite{li2020fourier}, Graph Neural Networks \cite{sanchez2020learning}, and recently attention based / transformer architectures, e.g., \cite{alkin2024universal, raonic2024convolutional, hao2023gnot, wu2023solving,luo2025transolver++}. 

\paragraph{Physics-Constrained Methods:} These methods use physics knowledge to constrain ML approaches. The most prominent are Physics Informed Neural Networks, where physics equations are introduced as constraints in the loss function \cite{lagaris1998artificial,raissi2019physics}. Adapting zero-shot learning corresponding methods could partially replace classical solvers. Extensions are multifold, e.g., using domain decomposition \cite{dolean2023multilevel} or reservoir-computing concepts \cite{chen2022bridging}, significantly enhancing their effectiveness. 

\subsection{Hybrid CFD Solvers}
\label{sec:LC_CNN_models}

Hybrid methods, focus of this paper, combine a classical solver based on a coarse mesh - the base solver - enriched with ML methods to achieve a comparable accuracy as obtained on a fine mesh \cite{sanderse2024scientific}, see Figure \ref{fig:overview}. Instead of computing an "exact" velocity $\vec{u}$, they resolve "only" a coarse-grained velocity $\vec{U}$ and introduce additional corrections / modifications to Equations (\ref{eq:NS01})-(\ref{eq:NS02}) ensuring that the effects of unresolved scales are appropriately reflected, i.e., $\vec{U}(t) = <\vec{u}(t)>$ with an appropriate averaging $<\cdot>$, maximizing accuracy and robustness. Unresolved scales could be recovered via super-resolution methods \cite{shu2023physics}, however in most industrial application one is interested only in integral quantities, such as drag and lift, which can be effectively calculated from coarse-grained velocities. Typically hybrid methods require significantly less training data than Operator Learning Methods \cite{melchers2023comparison}.

Here, we follow closely \citeauthor{um2020solver} (\citeyear{um2020solver}) and \citeauthor{kochkov2021machine} (\citeyear{kochkov2021machine})
which additionally introduce an autoregressive training approach. That is, during training sets of velocitiy trajectories $(\vec{U}_{i}, \vec{U}_{i+1}, \ldots \vec{U}_{i+N})$ are compared to the down-sampled outputs of a reference high resolution solver. The loss, measuring the mismatch, is then minimized through gradient descent. That is, the learned model is applied many times, autoregressively, for a single training step and thus a trajectory is fit and not only a single time step as in many earlier approaches \cite{melchers2023comparison}. Thus, the gradients need to "flow" through the solver \cite{list2024temporal} and corresponding methods are referred to \textit{solver-in-the-loop}.

Specifically, \citeauthor{um2020solver} (\citeyear{um2020solver}) introduces a learned correction (LC) approach, where a corrective term is added to the output of a coarse solver after each time step, i.e. to $\vec{U}_i$. The correction term is approximated by a neural network taking as input the current velocity field. The corrected velocity field is then used as the input velocity $\vec{U}_{i-1}$ in the subsequent time step. 

\citeauthor{kochkov2021machine} (\citeyear{kochkov2021machine}) also studied this learned correction approach and compared it to a learned interpolation (LI) approach where the coefficient for the interpolation of the velocity in the advection scheme of the solver are learned. More recently, \citeauthor{sun2023neuralpdesolvertemporal} (\citeyear{sun2023neuralpdesolvertemporal}) proposed to improve upon it by including temporal information as an input to the network. The LI approach can be thought as a way to incorporate more physical prior to the learned model because the learned coefficients can be constrained to be at least first order accurate. Compared to the LI approach, the LC approach is easier to integrate in an already existing industrial CFD code due to its minimally invasive architecture. Thus, for the rest of the paper we focus on the LC model. 

\begin{figure}[H]
    \centering
    \includegraphics[width=0.95\columnwidth]{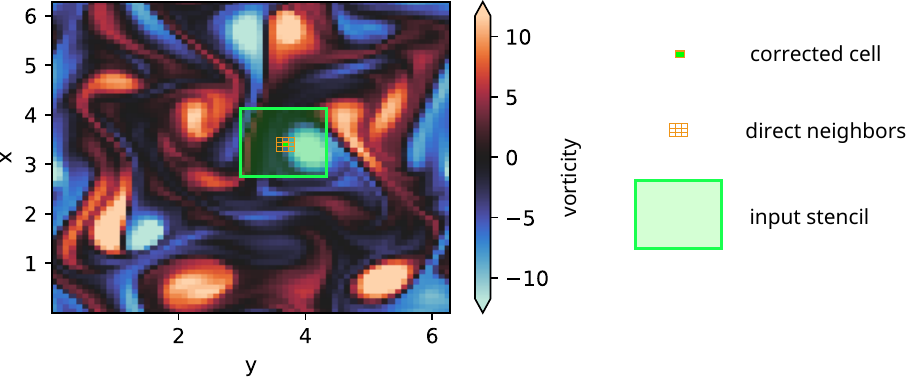}
    \caption{Input stencil for the corrector for a given cell on the forced turbulence : $64\times64$ grid on $[0;2\pi]\times[0;2\pi]$ domain. Case taken from \citeauthor{kochkov2021machine}.}
    \label{fig:input_stencil}
\end{figure}

In both \citeauthor{um2020solver} (\citeyear{um2020solver}) and \citeauthor{kochkov2021machine} (\citeyear{kochkov2021machine}) a CNN is used for the corrective term (as for the LI approach of \citeauthor{kochkov2021machine}). That is, for computing the correction to the velocity at a given cell the network is drawing information from a rather large stencil around the considered cell. For instance, the architecture used in \citeauthor{kochkov2021machine} consists of 7 layers of convolutions with $3\times3$ kernels, implying that the network uses information from a $15\times15$ stencil around the cell for which the correction is computed\footnote{In the first convolutional layer, a given neuron uses information from a $3\times3$ stencil around the cell, but at the next layer, the $3\times3$ kernel draws information from the neighbouring neurons which themselves contain information from a $3\times3$ stencil around their respective input cells. That is, the second layer sees information from a $5\times5$ stencil. At the 7\textsuperscript{th} and last layer the network finally has access to information coming from a $15\times15$ stencil around a given cell}, c.f., Figure \ref{fig:input_stencil}.

The non-local nature of the correction computed through CNNs brings significant benefits since many fluid properties, e.g., due to turbulence, usually depend on the history of flow pathlines. Our experiments have shown that the larger the non-locality of the correction, the more accurate predictions are (see Section \ref{sec:stencil_study}). That is, the performance of the learned correction approach is directly linked to the size of the CNN stencil from which the information is drawn. This underlines that CNN based  approaches are able to indirectly resolve the Lagrangian nature of flows within Eulerian formulations by going back in space instead of going back in time\footnote{That is, while the physics, and as such intrinsic properties as turbulent structures, are transported  with the fluid, any information is stored usually in a grid fixed in space and not moving with the fluid.}. In contrast, most classical turbulence models, which are formulated in terms of local interactions only, rely on additional transport equations, respectively transported variables, \cite{pope2000turbulent} to capture the Lagrangian nature.

\section{Transported Memory Networks}

\subsection{Motivation}

While using CNNs is straightforward and efficient for CFD solvers based on Cartesian grids, their application in industrial high-end CFD solvers remains impractical. Most solvers adopt a matrix-free approach, which efficiently accesses information only for nearest neighbors and a limited number of previous time steps. While replacing CNN with Graph Neural Networks may seem like a natural extension to handle unstructured meshes \cite{shankar2023differentiableturbulenceii}, this does not address the issue of having to access information beyond the immediate neighbours. 

Thus the question is: \textit{Is it possible, through a change in the ML model, to compute the learned correction by only using information coming from direct neighbours and still achieve the desirable performance?}

\subsection{Transported Memory Network Architecture}
\label{sec:TMN_arch}

Within this paper, we propose a new architecture, that is compatible with small computational stencils while still achieving the desired level of performance, both in term of accuracy and model efficiency. Inspired by various recurrent neural network architectures \cite{hochreiter1997long, gers2000recurrent, cho2014learning} our approach takes into account, in addition to the short term physical states $\vec{U}(t)$, a long term state (memory) encoded by a hidden state vector $\vec{H}(t)$, both defined cell / discretization point-wise (see Figure \ref{fig:overview} and \ref{fig:cell}). That is, the hidden states are latent variables that can be inferred only indirectly. Given the Lagrangian nature, the hidden state vector is effectively transported with the flow, as we show in Section \ref{sec:latent_discussion}. We thus refer to the architecture as Transported Memory Networks (TMN). It can be considered as a fusion of simulation algorithms formulated in a Eulerian coordinate system and recurrent neural network architectures.

Concretely, we use the FV approach on a staggered grid \cite{griebel1998numerical} based on the JAX-CFD code of \citeauthor{kochkov2021machine} (\citeyear{kochkov2021machine}), where in addition to the velocity components carried by each cell we add a hidden state vector defined at cell center that encodes history information. In practice, at each time step (with constant size $\Delta t = t_i - t_{i-1}$), a new intermediate velocity field is produced by the base solver, i.e., $\vec{U}^*_i =  \textbf{solver} ( \vec{U}_{i-1} )$, which is then corrected by the velocity corrector network using as input the current intermediate velocity (of the direct neighbours only) plus the hidden state vector (of the direct neighbours only), i.e., $\vec{U}_i = \vec{U}^*_i + \textbf{corr}_{\theta}(\vec{U}^*_i, \vec{H}_{i-1} )$. The hidden state vector is then itself updated, using its own update network taking as input the previous values and the current corrected velocity, i.e, $\vec{H}_{i} = \textbf{up}_{\theta} (\vec{U}_i, \vec{H}_{i-1} )$. Additionally we have also explored adding a physics prior by introducing an explicit transport of $\vec{H}$, but this has been less effective (see Appendix \ref{sec:stuff_tried}).

For initializing the hidden state vector we use a separate encoder network which computes the initial condition for the hidden state vector from the initial condition of the velocity field, i.e, $\vec{H}_0 = \textbf{enc}_{\theta}( \vec{U}_0 )$. Here, the encoder is non-local, with 7 layers of convolution with $3 \times 3$ kernel, meaning an "input stencil" of $15 \times 15$ as in the original model of \citeauthor{kochkov2021machine} (\citeyear{kochkov2021machine}). At encoding we do not have access to any previous snapshots in time and use the spatial information instead to encode the "history" of the flow\footnote{In classical turbulence model the additional quantities (like turbulence kinetic energy) also need to be initialized but this can be done based on their definition and the initial velocity field.}. The encoder is used only once at the beginning of the simulation, leaving the locality during solver iterations intact. 

The whole algorithm is summarized in Algorithm \ref{alg:LC_hidden states} (where the subscript $\theta$ denote the learnable functions) and further details of the architecture is given in Appendix \ref{sec:nets_architecture}. Albeit we assume a constant time step an extension to adaptive time steps, i.e., more general time integration schemes, is possible, e.g., \cite{chen2018neural,kang2023learning}.

\begin{algorithm}[t]
  \caption{Learned correction with hidden states}\label{alg:LC_hidden states} 
  \begin{algorithmic}
    \STATE Set initial conditions $\vec{U}_0$
    \STATE Encode initial hidden states $\vec{H}_0 = \textbf{enc}_{\theta}( \vec{U}_0 )$
    \WHILE{$1 \leq i \leq N$}    
      \STATE Update velocity with base solver $\vec{U}^*_i =  \textbf{solver} ( \vec{U}_{i-1} )$
      \STATE Correct velocity $\vec{U}_i = \vec{U}^*_i + \textbf{corr}_{\theta}(\vec{U}^*_i, \vec{H}_{i-1} )$
      \STATE Update hidden states $\vec{H}_{i} = \textbf{up}_{\theta} (\vec{U}_i, \vec{H}_{i-1} )$
    \ENDWHILE
  \end{algorithmic}
\end{algorithm} 


\section{Experiments}

\subsection{Setting}
\label{subsec:setting}

All models are trained with a base solver operating on a $64\times64$ grid with reference data coming from a $2048\times2048$ discretization. To facilitate comparability  of models, we strictly follow the data generation procedures presented in \citeauthor{kochkov2021machine}. 

The training data is the result of simulations of a 2D Kolmogorov flow in a box of side-length $2\pi$ on a structured grid with periodic boundary conditions. Furthermore, Reynolds number $\RN=1000$, density $\rho=1$, dynamic viscosity $\mu=0.001$, explicit time stepping with $\Delta t = 7.0125 \cdot 10^{-3}$ and forcing $f = \sin (4y) \vec{e}_1 - 0.1 \vec{U}$ (\textit{forced turbulence}) are used, where $\vec{e}_1$ is the basis vector in $x$-direction and $\vec{U}$ is the velocity vector. We generate 50 trajectories resulting from different random initial conditions. Of those, two are kept for the validation set and 16 for the test set. We also test our model on the three generalization cases from \citeauthor{kochkov2021machine} (\citeyear{kochkov2021machine}): a \textit{larger domain} with increased side-length of $4\pi$, a \textit{more turbulent} Kolmogorov flow with $\RN=4,000$, and a \textit{decaying turbulence} case where the forcing is removed.
 
\subsection{Training}
\label{subsec:training}

The networks are trained by consecutively minimizing the loss functions 
{\footnotesize
\begin{align}
 L_T = &\sum_{j=0}^{T/N} \mathsf{MSE}\left( \textbf{enc}_{\theta}(\vec{U}_{1 + jN}^{\text{exact}}), \vec{H}_{1 + jN} \right) 
    \notag\\
    &+ 
    \sum_{i=1}^T \mathsf{MSE}\left(\vec{U}_i^{\text{exact}}, \vec{U}_i\right)
    \label{eq:lossfn} \quad \text{for} \;\, T\in \{16,32,64\}
\end{align}}

\noindent
The second term is the standard mean square error between the down-sampled high resolution velocity field $\vec{U}_i^{\text{exact}}$ and the velocity field $\vec{U}_i$ obtained from the ML-augmented solver (see \citeauthor{kochkov2021machine} for more details on the down-sampling scheme). In practice, we found that starting the training on trajectories composed of 16 time steps, and then increasing its length to 32 and finally to 64 time steps worked best for the TMN approach. These chunks are obtained by splitting up longer trajectories of $4,800$ sequential time steps. 

The first term computes the difference between the hidden state vector after $N$ time steps and the output of the encoder taking as input the velocity at this same $N$\textsuperscript{th} time step (of the chunk of trajectory). $\textbf{enc}_{\theta}(\vec{U}_{1 + jN}^{\text{exact}})$ denotes the output of encoder when taking as input the velocity (of the reference model\footnote{We also explored using the velocity produced by the corrected model but did not observe any difference in model performance.}) at time $t_{1 + jN}$, and $\vec{H}_{1 + jN}$ is the hidden state vector at the same time step, i.e., after $jN$ hidden state updates of the initial encoded value $\textbf{enc}_{\theta}(\vec{U}_{1}^{\text{exact}})$. This term ensures that the hidden state that is encoded at the end of a chunk will be similar to the one that is encoded at the beginning of the next chunk (in the same trajectory) and thus enforce stability during long roll-outs.

The initial learning rate of 0.001 is exponentially decreased with a decay rate of 0.5 and 50,000 transition steps. A batch size of 8 chunks is used. At each chunk length, 400 training epochs are performed producing converging loss curves. 

The error is computed on trajectories, instead of single time steps, which means the correction is applied auto-regressively, and the gradients used in the optimization procedure take into account the effect of the correction not only on the current time step but also on all the subsequent ones in the trajectories, see \citeauthor{melchers2023comparison} for an in-depth explanation. 
\subsection{Evaluation Metrics}
\label{subsec:metric}

To compare the performances of the models, we use the same metrics as in \citeauthor{kochkov2021machine}. As a metric for point-wise accuracy, we evaluate the Pearson correlation of the vorticity field $\nabla\times\vec{U}$ produced by the models with the one obtained from reference high resolution solvers. To facilitate comparison across models, we look in particular at the time at which the correlation falls below $0.95$. 

For statistical accuracy, we compute the energy spectrum $E(k) = 0.5 |\vec{U}(k)|^2$, where $k$ is the wavenumber, and evaluate the spectral error defined as the average absolute error of $E(k)k^5$ between the reference solution and the respective model output.

For all metrics, the numbers reported are taken from an average over 16 different trajectories. We also tested the robustness of each model by varying the random initialization of the parameters of the neural networks with 4 different random seeds. As baselines we report the performances of uncorrected base solvers on intermediates grid sizes.

\subsection{Effect of the Input Stencil Size}
\label{sec:stencil_study}

As mentioned in Section \ref{sec:LC_CNN_models}, the original LC model in \citeauthor{kochkov2021machine} (\citeyear{kochkov2021machine}) uses information coming from a $15\times15$ stencil around a given cell to compute the correction for that same cell. To identify the influence of the stencil size on model performance, we track the accuracy of a series of models with varying stencil size. The models consist of a variable number of convolution layers with $3\times3$ stencils, followed by two layers with $1\times1$ kernel that act as final, local Multilayer Perceptron (MLP). The stencil size is then equal to $2N + 1$ where $N$ is the number of convolution layers with $3\times3$ kernels (see Appendix \ref{sec:nets_architecture} for details).

\begin{figure}[h!]
         \centering
         \includegraphics[width=0.9\columnwidth]{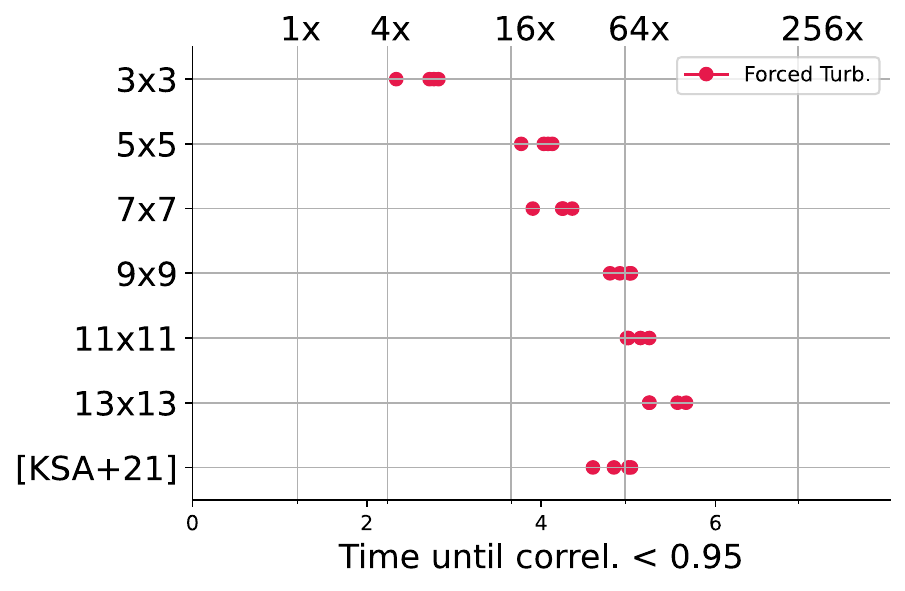}
     \caption{Simulation time until correlation drops below 0.95 for models with different input stencil sizes and 4 random initializations, vertical lines indicate the multiples of cells required by base solver to achieve the respective timings.}
     \label{fig:stencil_study}
\end{figure}

The time until the correlation falls below 0.95 for the various stencil sizes are reported in Figure \ref{fig:stencil_study} for the \textit{forced turbulence} case, with the performance of the model from \citeauthor{kochkov2021machine} added as reference. As expected the model performance improves with larger stencil, with a plateau achieved for stencils above $13\times13$.

In particular, the model with a $3\times3$ stencil (i.e. the one using only information coming from the direct neighbours) shows considerably worse performance compared to the best performing models. 

\subsection{Benchmarking the TMN approach}
\label{sec:latent_study}

\begin{figure}[h!]
    \centering
    \includegraphics[width=0.9\columnwidth]{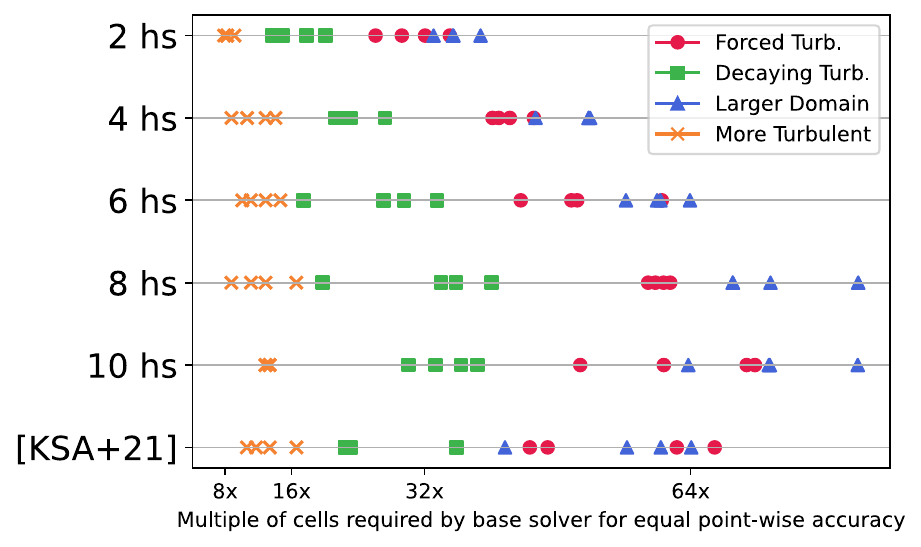}
    \caption{Simulation time until correlation drops below 0.95 for models with number of hidden states (hs) and 4 random initializations. For more details on the simulation settings, c.f., Section \ref{subsec:setting}. }
    \label{fig:results_vort_corr}  
\end{figure}

\begin{figure}[h!]
    \centering
         \includegraphics[width=0.9\columnwidth]{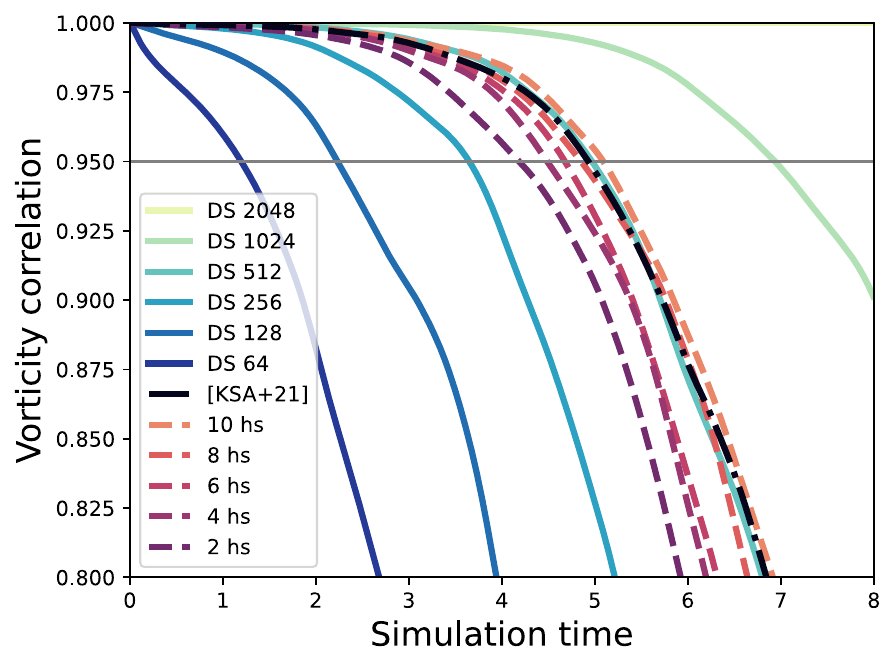}
    \caption{Simulation time until correlation drops below 0.95 for for direct simulations (DS) with different discretizations as well as models with varying number of hidden states (hs) for \textit{forced turbulence}.}
    \label{fig:corr_time_force_turb}
\end{figure}

\paragraph{Accuracy:} 
In Figure \ref{fig:results_vort_corr}  we report the results of the TMN model for different sizes of the hidden state vector and for the various cases: \textit{forced turbulence} (same setting as the training data), \textit{decaying turbulence}, \textit{larger domain} and \textit{more turbulent} (see Section \ref{subsec:setting} for more details). Figure \ref{fig:corr_time_force_turb} shows the correlations over time for the \textit{forced turbulence} case.

While on average the model accuracy does improve with the number of hidden states, the trend is not as pronounced as in the case of the stencil size (see Section \ref{sec:stencil_study}). Overall, a hidden state vector of dimension 8 is enough to obtain comparable results to the full CNN architecture, but even models with as few as 4 hidden states remain competitive in most cases.

\begin{figure}[h!]
     \centering
     \begin{subfigure}[b]{0.49\columnwidth}
         \centering
         \includegraphics[width=0.99\columnwidth]{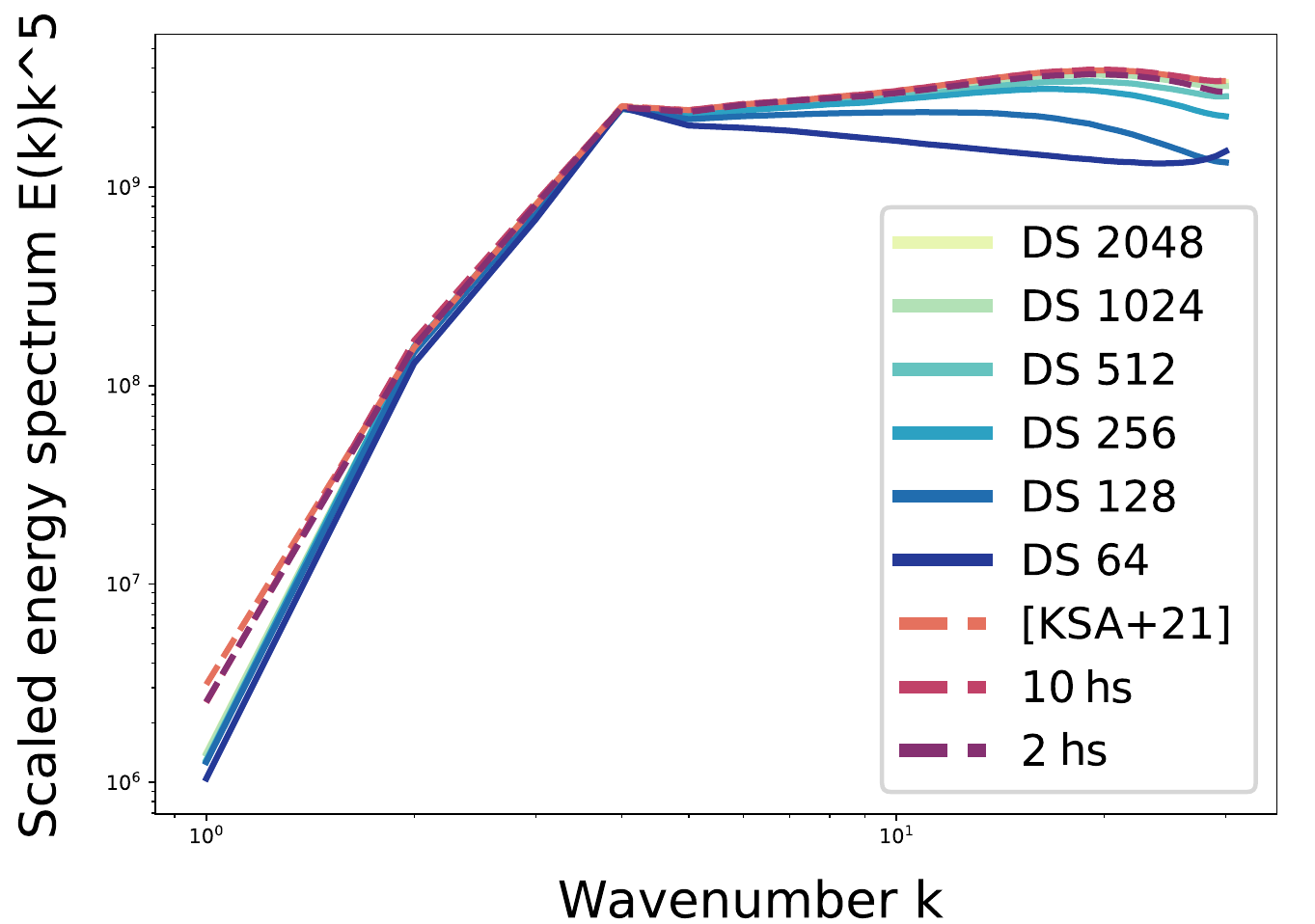}
     \end{subfigure}
     \hfill
     \begin{subfigure}[b]{0.49\columnwidth}
         \centering
         \includegraphics[width=0.99\columnwidth]{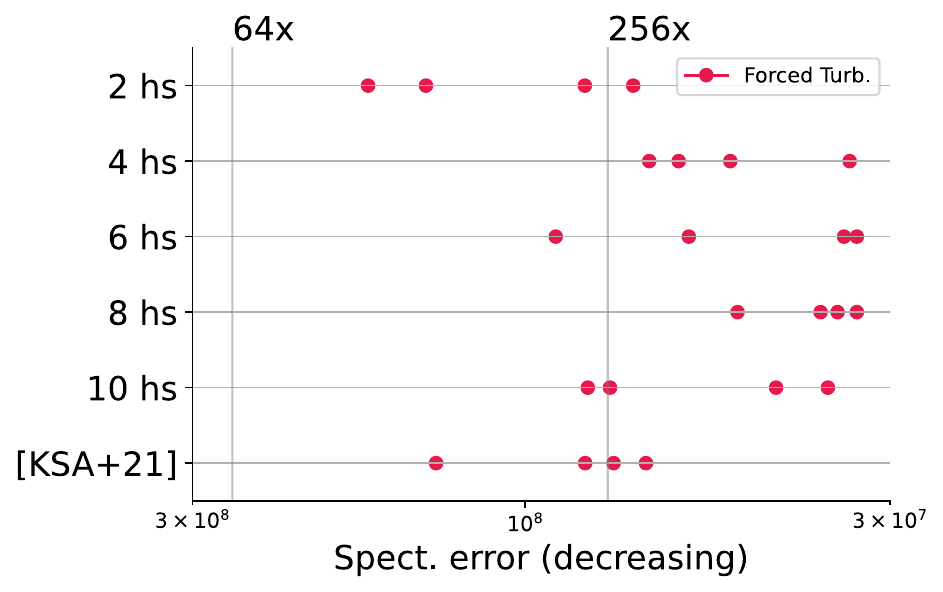}
     \end{subfigure}
     \caption{Left, the energy spectrum scaled by $k^5$ over wavelength is shown for direct simulations (DS) and learned models for \textit{forced turbulence}. Right, the spectral error is depicted for learned models (dots) for different random initializations, vertical lines indicate multiple of required cells by the base solver,}
     \label{fig:energy_forc_turb}
\end{figure}

Figure \ref{fig:energy_forc_turb} shows the energy spectrum (scaled by $k^5$ where $k$ is the wave number) for the various baselines, for the TMN models with 2 and 10 hidden states and for the model from \citeauthor{kochkov2021machine} (\citeyear{kochkov2021machine}). The learned models agree well with the high resolution baseline solution except at $k=1$. Further, Figure \ref{fig:energy_forc_turb} depicts the spectral error (see Section \ref{subsec:metric}) for the 2 to 10 hidden states TMN models and 4 different random initialization each. When satisfied with statistical accuracy, 2 hidden states is enough to reach a comparable accuracy as the original LC model in \citeauthor{kochkov2021machine} (\citeyear{kochkov2021machine}). Using more than 4 hidden states does not seem to bring further improvement for this metric.

\paragraph{Speed:} 
In Table \ref{tab:models_speed} we report the runtime per time unit (i.e., how long does it take to compute one second worth of simulated flow) as a function of the number of hidden states, with the model from \citeauthor{kochkov2021machine} added for reference. All models were run on a single Nvidia A10G GPU.


\begin{table} [h!] 
\small
    \centering
        \begin{tabular}{@{\,}ccccccc|c@{\,}} 
        \toprule
        \# Hidden states &  2 & 4 & 6 & 8 & 10 & 12 & \citeauthor{kochkov2021machine}\\
        \hline 
        Speedup   &  
        99 & 95 & 93 & 90 & 88 & 85 & 86 \\
        \bottomrule
    \end{tabular}
    \caption{Speedup of simulation for different sizes of the hidden state vector. The simulation times of the corresponding hybrid methods on a $64\times64$ grid are compared with the direct simulation on a $512\times512$ grid, which has a comparable accuracy (i.e., accuracy correlation as shown in Figure \ref{fig:corr_time_force_turb}).}
    \label{tab:models_speed}
\end{table}



These timings demonstrate that the TMN architecture is computationally faster for all models up to 10 hidden states compared with the CNN model from \citeauthor{kochkov2021machine}. However, we would like to point out that we do not perform any optimization regarding the latency of the TMN models. Thus potential further speed up can be expected with more carefully chosen implementations and network architectures. Last but not least, we expect additional speedup going from 2D to 3D due to cubic scaling of compute effort in 3D versus quadratic scaling in 2D.

We also note that, as reported in \cite{mcgreivy2024weak}, in the original study of \citeauthor{kochkov2021machine} the LC and LI models were only evaluated against an (uncorrected) FV solvers, and not the stronger baselines of Pseudo-Spectral \cite{shen2011spectral} and DG methods \cite{cockburn2012discontinuous}, which would perform significantly better in the benchmarks selected here. However, in our case we are concerned with industrial solvers which primarily use FV methods. Thus it is a fair baseline.

\subsection{Transport of Hidden State Vector}
\label{sec:latent_discussion}

\begin{figure}[h!]
    \centering
    \includegraphics[width=0.9\columnwidth]{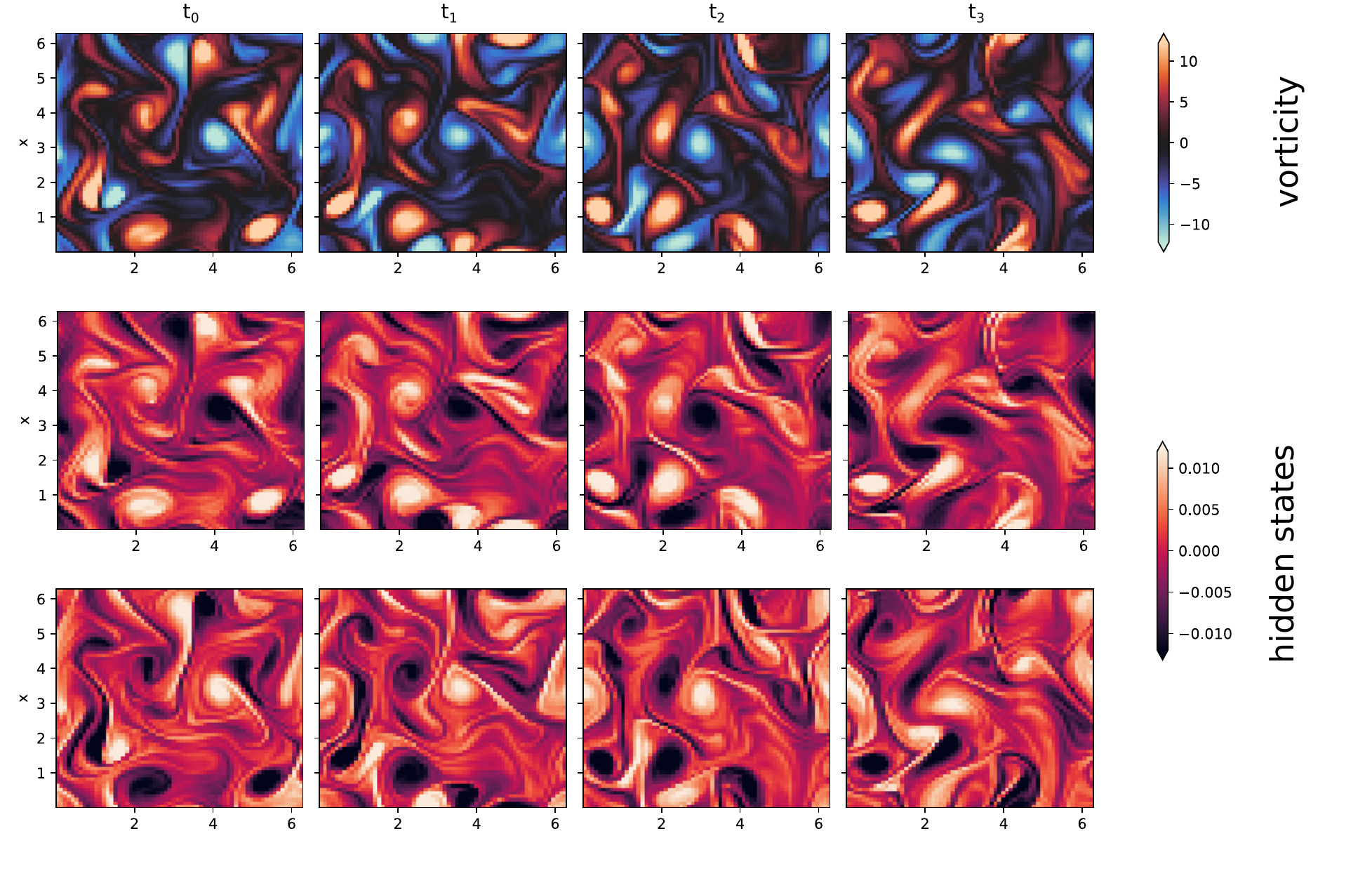}
    \caption{Vorticity (row 1) and first 2 hidden states (rows 2-3) show remarkable resemblance.}
    \label{fig:vort_and_lat}
\end{figure}

Some interesting insight can be gained by inspecting the values and evolution of the hidden states during a trajectory. Figure \ref{fig:vort_and_lat} show the evolution of the first two components of the long term state $\vec{H}$ for one of the trajectories of the test set in the forced turbulence case. Also, the vorticity of the velocity field is pictured. 


As can be seen from the plots, the hidden state components are highly correlated with the vorticity. This suggest that the update
{\footnotesize
\begin{align}
    \vec{H}_{i} = \textbf{up}_{\theta} (\vec{U}_i, \vec{H}_{i-1} )
    \label{eq:latent_update}
\end{align}}

\noindent 
effectively learns a transport equation (the vorticity itself is transported with the flow and any transported quantity would follow a similar pattern). We stress that during training the model never sees the vorticity directly, but only the velocity components, see Equation (\ref{eq:lossfn}).

As an alternative to the direct update of Equation (\ref{eq:latent_update}), and taking inspiration from classical turbulence model, we tried to explicitly transport the hidden states and only learn a source term with different variation of the update:
{\footnotesize
\begin{align}
    \vec{H}_{i} = \textbf{trsprt}\left(\vec{U}_i, \vec{H}_{i-1}\right) + \textbf{s}_{\theta} (\vec{U}_i, \vec{H}_{i-1} )
    \label{eq:latent_transport}
\end{align}
}

\noindent
with $\textbf{trsprt}$ the transport of quantity $\vec{H}_{i-1}$ by the velocity field $\vec{U}_i$ given by the base solver and $\textbf{s}_{\theta}$ a learnable source component. However this approach with a physics prior by explicitly modelling transport negatively impacts performance (see 
Appendix \ref{sec:stuff_tried} for more details).

\section{Conclusion and Perspective}

In this paper we propose a novel architecture for the \textit{solver-in-the-loop} approach originally introduced in \citeauthor{um2020solver} (\citeyear{um2020solver}) and \citeauthor{kochkov2021machine} (\citeyear{kochkov2021machine}). Both papers relied on a CNN to correct the outputs of a standard numerical base solver for the time dependent Navier-Stokes equations. We have shown here that the performance of the CNN based architecture is highly reliant on a large input stencil (the size of the window from which the information is drawn) to compute its correction. While this is not an issue for Cartesian grid based solvers, it makes the approach hard to integrate in existing large-scale industrial solvers working on unstructured meshes. In the latter case,  the "only" information available per cell are the variables (velocity, pressure, etc.) coming from the direct neighbouring cells. As an alternative to CNN based architecture, we propose a model compatible with simple feed forward networks (shared by each cell) and using information from direct neighbours only. Since information from cells farther away is not directly available, we use an additional hidden state vector, stored cell-wise, to encode missing information. Crucially, this hidden state vector needs to be updated between each time step and we do so in a fully learnable way, based on the current velocity. The hidden state vector thus replaces the spatial information available to the CNN based architecture with an history dependent one. A parallel can be drawn to the additional physical variables that are transported with the flow in classical turbulence models. Indeed, analysis of the hidden state vector during the flow history shows that it is also transported, even if the architecture itself does not explicitly enforce it. We call this approach Transported Memory Network (TMN). Compared to classical CNNs the approach relies only on direct neighbor information and thus is relative geometric agnostic. This will likely improve generalization capabilities in realistic industrial scenarios. 

Our results show that the TMN architecture compares very well with the CNN based architecture by \citeauthor{kochkov2021machine} (\citeyear{kochkov2021machine}), yet shows advantages in terms of speed (latency of the model). To obtain comparable point-wise accuracy 8 hidden states are sufficient and for equivalent statistical accuracy only 2 hidden states are required. Since furthermore only local input is required, the TMN architecture is thus a strong candidate to bring the \textit{solver-in-the-loop} approach in state-of-the-art industrial CFD codes. 

The next steps to be addressed towards full integration include testing the approach on actual unstructured meshes (including in 3D), handling of more general boundary conditions, and making it compatible with variable time steps. Last but not least we plan to do some more excessive benchmarking as well as ablation studies.

\appendix
\section{Networks Architectures}\label{sec:nets_architecture}
\subsection{CNNs with Varying Stencil Size}

\begin{figure}[hbt!]
   \centering
   \includegraphics[width=0.40\columnwidth,angle=90]{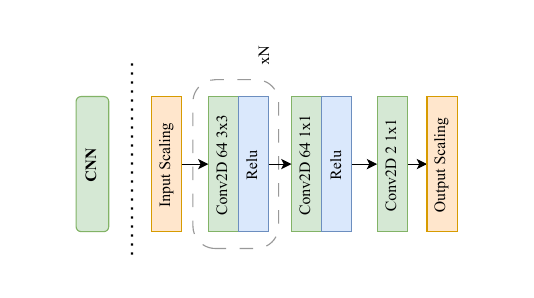} 
   \caption{Architectures for the stencil size study}
   \label{fig:stencil_arch}
\end{figure}

Figure \ref{fig:stencil_arch} shows a diagram of the series of CNN models used for the study of the effect of the input stencil on the corrector performance in Section \ref{sec:stencil_study}. The only variable parameter between models is $N$, the number of convolution layers with $3\times3$ kernels, going from $N = 6$ for the $13 \times 13$ input stencil to $N = 1$ for the $3 \times 3$ input stencil. The number of trainable parameters for each networks range from $190,146$ to $5,506$.

\subsection{Transported Memory Network}
The TMN model is comprised of three different learnable components that work on top of a base solver (see Algorithm \ref{alg:LC_hidden states}). The hidden state encoder generates the initial condition for the hidden state vector based on the initial velocity field. The velocity corrector corrects the velocity based on its current value and the hidden state vector, and finally the hidden state updater updates the hidden state vector to the next time step based on its current value and the corrected velocity.

\begin{figure}[h!]
    \centering
    \includegraphics[width=0.99\columnwidth]{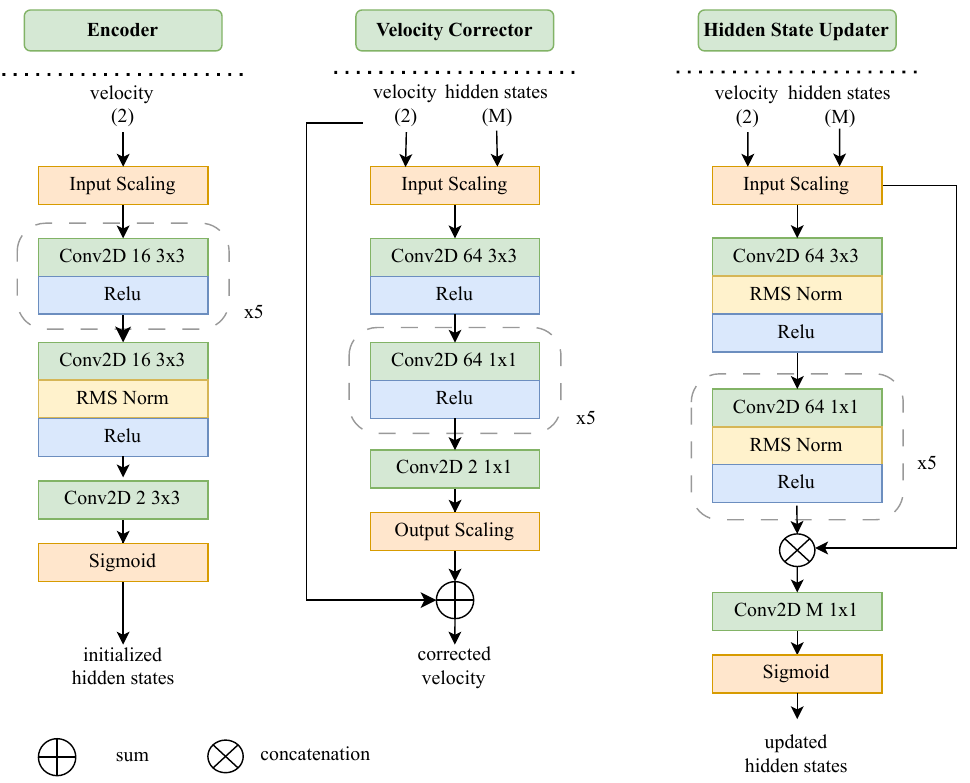} 
    \caption{Local Networks architectures, size of the velocity (2) and hidden state vector (M) are indicated in brackets.}
    \label{fig:nets_architecture}
\end{figure}

The architecture of the three components are depicted in Figure \ref{fig:nets_architecture}. Note that for both the velocity corrector and the hidden state updater, the first convolution layer employs a $3\times3$ kernel while all subsequent ones use $1\times1$ kernels. Even though those networks are implemented using CNNs, they are equivalent to a single MLP sliding over the grid with an input stencil of $3\times3$ (i.e., only using direct neighbors information as input). 
The sigmoid used as final output function for both the encoder and the hidden state updater bounds the hidden state vector components within $[-1,1]$.

Note that for the hidden state updater the velocity and hidden state vector field are concatenated to the intermediate output before the final layer rather than added after the final layer. This ensures that the values for the hidden state vector always remains within $[-1,1]$. We found that the alternative additive updater reduces the numerical stability, especially over long roll-outs.

Depending on the size of the hidden state vector, the total number of parameters for the 3 trainable networks range between $56,210$ (12 hidden states) to $42,430$ (2 hidden states).

\section{Further Experiments}
\label{sec:stuff_tried}

While the overall structure of the learned model with hidden states remained the same throughout our study  (i.e., with 3 separate components: hidden state encoder, velocity corrector, and hidden state updater) several variations of the individual components were explored. We detail here some of those variations that were however not retained for the final study due to inferior performance.

\paragraph{Explicit Transport of the Hidden State:}
As mentioned in Sections \ref{sec:TMN_arch} and \ref{sec:latent_discussion}, we are using the fully learnable update rule of Equation (\ref{eq:latent_update}). Additionally, we also explored alternatives with more inductive bias explicitly transporting the hidden states with the velocity as in Equation (\ref{eq:latent_transport}). Specifically we tried (with $H^i$ the i-th component of $\Vec{H}$):
{\footnotesize
\begin{align*}
    \textbf{trsprt}(l^i) = - \nabla \cdot (H^i \Vec{U}) +  \nabla \cdot ( \xi (\Vec{U}, \Vec{H}) \nabla H^i)
\end{align*}}
with 
\begin{itemize}
    \item advection only, i.e. $\xi \equiv 0$,
    \item advection-diffusion with learnable constant viscosity $\xi$,
    \item advection-diffusion with learnable non constant viscosity  with $\xi (\Vec{U}, \Vec{H}) $ parametrized by a simple neural network.
\end{itemize}
However, we did not observe a performance improvement adding a physics prior.

\paragraph{Recurrent Architectures:} The transported memory network can be viewed as a type of recurrent architecture with the hidden state vector encoding history information from the previous state in the sequence (the observed sequence here being the velocities components on a given cell at each time  step). We therefore explored an update mechanism based on the principles of the Gated Recurrent Unit (GRU) architectures \cite{cho2014learning} i.e. by updating the hidden state vector / hidden state through a combination of an update gate and a forget gate. We also tried the principles behind the Long Short Term Memory (LSTM) networks \cite{hochreiter1997long}, splitting the hidden state vector into an internal state designed to carry long term history and an hidden state for short term memory, updated through specific output gates, forget gates and input gates. Introducing corresponding slightly more complex architectures, we could not observe a performance improvement.

\bibliography{bibliography}

\end{document}